\documentclass[10pt,twocolumn,letterpaper]{article}

\usepackage{cvpr}              
\definecolor{cvprblue}{rgb}{0.21,0.49,0.74}
\usepackage[pagebackref,breaklinks,colorlinks,allcolors=cvprblue]{hyperref}
\usepackage{algorithm}
\usepackage{algorithmic}
\usepackage{amsmath}
\usepackage{bm}
\usepackage{amssymb}

\usepackage{colortbl}  
\definecolor{Green}{RGB}{240, 255, 255}
\definecolor{Gray}{RGB}{237, 252, 214}

\usepackage{multirow}

\def\paperID{*****} 
\def\confName{CVPR}
\def\confYear{2026}

\title{Ultra: Unsupervised Cross-Task Optimization for Reliable Restoration Segmentation Collaboration under Adverse Weather}

\newcommand{\whu}{1}%
\newcommand{\bit}{2}%
\newcommand{\thu}{3}%
\newcommand{\bza}{4}%
\author{
Shiqin Wang\textsuperscript{\whu}\quad
Zhiqian Li\textsuperscript{\whu}\quad
Haoyuan Du\textsuperscript{\whu} \quad
Junming Chen\textsuperscript{\whu}  \quad 
Jiayuan Li \textsuperscript{\bit, \bza} \quad \\
Tianrun Xu \textsuperscript{\thu, \bza}\quad
Haoyang Chen \textsuperscript{\whu,\bza}\quad\quad \\
\normalsize{\textsuperscript{\whu}Wuhan University} \quad 
\normalsize{\textsuperscript{\bit}Beijing Institute of Technology} \quad
\normalsize{\textsuperscript{\thu}Tsinghua University} \quad
\normalsize{\textsuperscript{\bza}Zhongguancun Academy} \quad \\
{\tt\small shiqinwang@whu.edu.cn, haoyangchen@whu.edu.cn}
}

\begin{document}
\maketitle
\begin{abstract}
Unsupervised Domain Adaptation for Adverse Weather Semantic Segmentation (UDA-ASS) aims to transfer semantic knowledge from labeled normal-weather images to unlabeled adverse environments. Existing approaches implicitly assume that restoration and segmentation provide mutually beneficial guidance. However, under severe degradation and without target-domain supervision, the validity of cross-task optimization directions becomes fundamentally unidentifiable, leading to hallucination-driven error propagation. In this work, we propose a novel Unsupervised Restoration-Segmentation Collaborative Learning Framework (Ultra), which reframes cross-task interaction as direction selection under uncertainty and causal effect estimation, enabling reliable collaboration through candidate direction generation and intervention-based filtering. In detail, we propose CTDN and CMIL. The former exploits complementary visual structures and semantic information to generate candidate optimization directions and performs cooperative direction selection between restoration and segmentation. The latter reformulates cross-task information transfer from correlation-based propagation into causal effect assessment, suppressing hallucination propagation. Extensive experiments on three widely used UDA-ASS benchmarks demonstrate state-of-the-art segmentation performance. Beyond segmentation, our framework achieves better unsupervised restoration results than existing UDA-ASS restoration methods and generalizes to unsupervised restoration and object detection collaboration tasks. Code and models will be available at https://github.com/Wang-Shiqin/Ultra.
\end{abstract}    
\section{Introduction}
\label{sec:intro}

Adverse weather Semantic Segmentation (ASS) aims to assign each pixel in adverse weather images, including rain, fog, snow, and low illumination conditions, to a corresponding object category. It has been widely applied in autonomous driving~\cite{li2024parsing,wang2025parables}, visual surveillance~\cite{yang2025semantic,xu2021exploring,wang2024low}, and robotic perception~\cite{wu2026beyondsparse}. 
While real-world adverse weather images can be collected easily, building matching datasets with precise pixel-level annotations takes enormous time and effort. This greatly restricts the advancement of traditional supervised learning methods for ASS. 
Some studies attempt to transfer semantic segmentation techniques developed under clear daytime conditions~\cite{wei2024stronger,yun2025soma,zang2026geco,zhang2026causal}. 
However, the substantial domain gap between clear-weather and adverse-weather scenarios leads to severe performance degradation in challenging weather environments. To reduce annotation requirements and bridge the domain gap, Unsupervised Domain Adaptation for Adverse Weather Semantic Segmentation (UDA-ASS) has attracted increasing attention by transferring knowledge from labeled daytime source domains to unlabeled adverse-weather target domains.

\begin{figure}[!t]
\centering
\includegraphics[width=0.96\columnwidth]{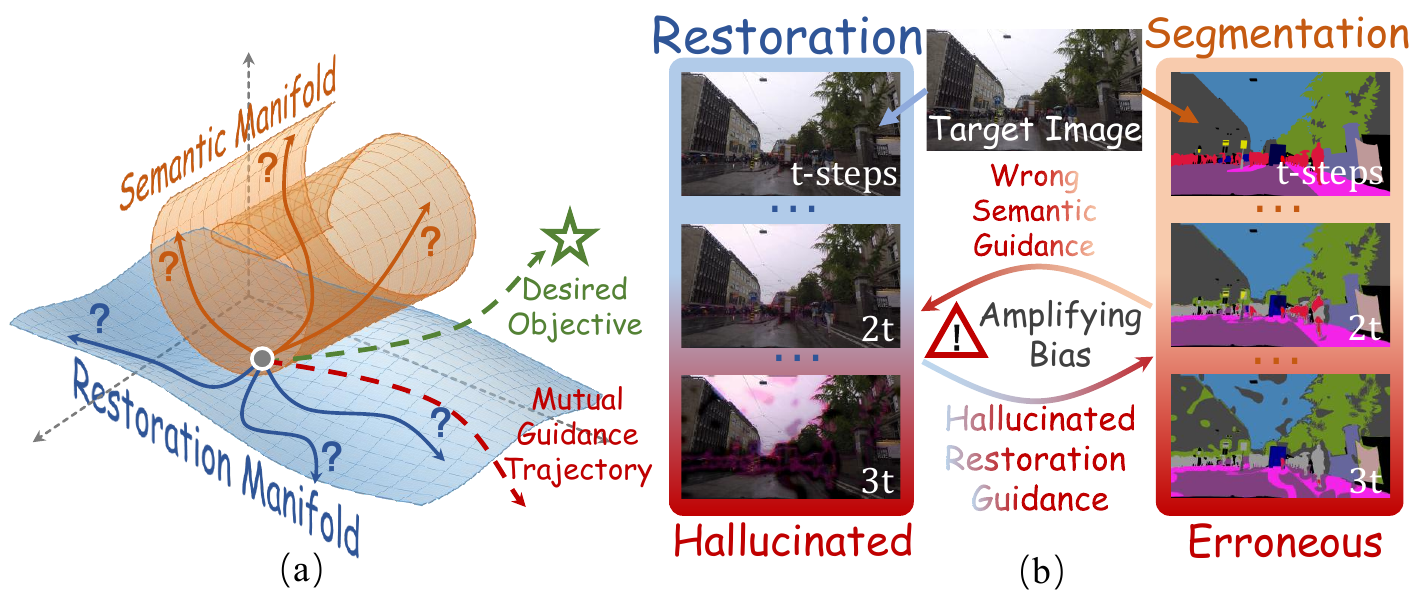}
\caption{ Challenges in unsupervised restoration-segmentation collaboration: (a) \textit{Ambiguous cross-task optimization directions} due to the lack of reliable supervision in heterogeneous task spaces; (b) \textit{Self-reinforcing hallucination loops} caused by iterative propagation of unreliable cross-task biases.}
\label{fig:problem}
\end{figure}

Existing UDA-ASS methods mainly follow two technical paradigms. 
The first paradigm focuses on enhancing cross-domain consistency learning to facilitate domain adaptation~\cite{bruggemann2022refign,bruggemann2023contrastive,li2024parsing,liu2024domain,pan2025exploring,xu2026crope,tian2026scm,chen2025mpsd}. However, enhancing prediction consistency makes the model overly rely on daytime priors and cannot guarantee that the predictions are truly supported by the visual evidence from adverse-weather images. 
The second paradigm reduces the visual discrepancy between source and target domains through style translation~\cite{zhengl2023compuda,shen2025w} or image/feature restoration~\cite{li2023vblc,lee2025frest,ying2025scdf}. 
Nevertheless, such restoration processes typically rely on natural image statistics as priors to compensate for missing information, which may introduce visually plausible but semantically nonexistent textures and incorrect structures, thereby misleading semantic reasoning and causing high-confidence hallucinated predictions.

These limitations suggest that restoration in UDA-ASS should not be treated as a task-agnostic preprocessing step: whether recovered observations are useful is determined by whether they support reliable semantic decisions. Once restoration is tied to semantic prediction, the two tasks become interdependent. Segmentation can discourage appearance-oriented recovery, while restoration can strengthen the evidence available to segmentation when target-domain cues are weak.

However, unlike previous work on supervised learning~\cite{guan2026restoration}, directly performing mutual guidance between restoration and segmentation under the unsupervised setting faces two critical challenges. \textit{(1) Ambiguous Cross-task Optimization Direction}. Since UDA tasks simultaneously lack clean image supervision required by restoration and semantic annotations required by segmentation, both tasks cannot determine whether their current optimization directions are reliable, resulting in adaptation trajectories that may deviate from the desired objectives during mutual guidance, as shown in Figure~\ref{fig:problem} (a). \textit{(2) Self-reinforcing Hallucination Loop.} Severe visual degradation leads restoration and segmentation models to rely on unreliable priors, resulting in hallucinated restoration structures and erroneous semantic predictions. During mutual interaction, these unreliable task-specific biases are repeatedly propagated and amplified, forming a closed-loop error accumulation process, as shown in Figure~\ref{fig:problem} (b).

To address these challenges, we propose a novel \textbf{Unsupervised Restoration-Segmentation Collaborative Learning Framework (Ultra)} that enables reliable cross-task cooperation by jointly discovering cooperative optimization directions and regulating task interactions.  
In detail, we propose \textbf{Cross-Task Direction Negotiation (CTDN)} for identifying cooperative optimization trajectories. Specifically, we first transform visual structural priors and semantic discriminative information into mutually constrained candidate update directions via bidirectional task adaptation between restoration and segmentation. Then, restoration and segmentation are modeled as cooperative agents with distinct optimization utilities. We construct a Nash bargaining process within the candidate update direction space to negotiate optimal update directions. These directions maximize the joint utility while balancing improved restoration quality and enhanced semantic discrimination.
To alleviate cross-task hallucination propagation caused by severe visual evidence degradation, we further propose \textbf{Causal Mutual Intervention Learning (CMIL)} module for preventing unreliable knowledge propagation between restoration and segmentation. In detail, we reformulate cross-task information propagation as task-level intervention effect estimation. By modeling the state evolution of a target task after the intervention of source-task factors, CMIL identifies cross-task intervention directions with positive optimization effects and suppresses negative information propagation caused by restoration hallucination and segmentation hallucination, thereby achieving reliable restoration-segmentation collaborative adaptation.

Extensive experiments on three widely used UDA-ASS benchmarks, together with image restoration evaluations in the UDA-ASS setting, demonstrate state-of-the-art segmentation and restoration performance.
Besides, the proposed framework is not restricted to specific task structures. Instead, it provides a general unsupervised complementary-task collaboration paradigm by learning reliable guidance relationships between different tasks. Therefore, the framework can be naturally extended to other task combinations, such as restoration-object detection collaboration. 
When extended to UDA adverse-weather object detection, our framework consistently improves existing UDA detection methods by enabling restoration-detection collaboration, demonstrating its effectiveness and generalization capability.

The contributions of this work are summarized as follows:

\begin{itemize}

\item We identify and formulate the problem of unsupervised restoration-segmentation collaboration under adverse weather adaptation, where the absence of target-domain restoration and semantic supervision causes two fundamental challenges: unreliable cross-task optimization directions and hallucination-induced error accumulation during mutual guidance. 

\item We propose Ultra, an Unsupervised Restoration-Segmentation Collaborative Learning Framework. It contains two core components: CTDN and CMIL. CTDN generates complementary candidate directions via cross-task interaction and uses Nash bargaining to select the optimal cross-task optimization direction. CMIL models cross-task knowledge exchange as task-level causal interventions to reduce error propagation caused by hallucinations.

\item Extensive experiments on three widely used UDA-ASS benchmarks demonstrate that our Ultra achieves state-of-the-art semantic segmentation performance. Moreover, Ultra significantly improves the quality of unsupervised image restoration over existing UDA-ASS methods. When applied to existing UDA adverse-weather object detection frameworks, our method achieves consistent performance improvements, validating its effectiveness and generalization ability.
\end{itemize}
\section{Related Works}
\label{sec:relatedwork}

\subsection{Adverse Weather Semantic Segmentation}
This task performs pixel-wise semantic parsing on images captured under degraded conditions (\textit{e.g.,} fog, night, snow, and rain).
Because collecting and annotating adverse weather data is substantially more costly than for clear daytime scenes, one group of studies resorts to clear-to-adverse domain generalization, where a segmentation model trained solely on a clear source domain is expected to generalize to arbitrary unseen adverse scenes~\cite{wei2024stronger,Bi_2025_ICCV,bi2025nightadapter,zhao2025fishertune,yun2025soma,zang2026geco,ye2026rfglnet}. Nevertheless, without observing adverse-weather data during training, these models still suffer considerable accuracy loss on adverse scenes. UDA-ASS thus serves as a more practical alternative, which exploits unlabeled adverse weather images and adapts the semantic knowledge learned in the labeled clear domain to the target domain. Research on UDA-ASS has evolved along two directions. The first direction pursues cross-domain consistency, either via learning more consistent feature representations~\cite{bruggemann2022refign,bruggemann2023contrastive,li2024parsing, liu2024domain, xu2026crope,tian2026scm,Wang_2026_CVPR} or via enforcing prediction consistency across domains~\cite{chen2025mpsd}; although such consistency benefits adaptation robustness, whether the resulting target predictions are grounded in reliable visual evidence remains unverified. The second direction narrows the domain shift at the appearance level, adopting weather style translation~\cite{zhengl2023compuda,shen2025w} or image/feature restoration~\cite{li2023vblc,lee2025frest,ying2025scdf} to pull adverse-domain representations closer to the clear domain. Since both directions concentrate on aligning distributions or enhancing appearance, they remain prone to segmentation hallucinations once visual evidence is severely degraded or missing.

\subsection{Task-aware Image Restoration and Segmentation}
Task-oriented Image Restoration methods aim to enhance downstream task performance, DIP dynamically adapts image processing according to degradation factors~\cite{liu2022image}. VDR-IR unifies semantic representations of diverse degraded images to effectively recover intrinsic semantics~\cite{yang2023visual}. Complementary to these approaches, another line of Segmentation-Guided Restoration works investigates how semantic information from segmentation models can guide image restoration~\cite{wu2024towards,zeng2025low}. However, existing methods mainly focus on unidirectional information transfer between image restoration and semantic segmentation~\cite{yang2023visual,yang2022self}. Consequently, the restoration or segmentation results heavily depend on the reliability of the restoration or segmentation process, leading to suboptimal restoration or segmentation performance.
To this end, Guan~\textit{et al.} first proposed a fully-supervised Restoration Adaptation for Semantic Segmentation (RASS) method, which enables joint optimization and mutual promotion between image restoration and semantic segmentation~\cite{guan2026restoration}. However, this fully-supervised paradigm requires paired restoration-segmentation supervision, which is unavailable in unsupervised adverse-domain adaptation scenarios.

\section{Methodology}
\label{sec:methodogy}

\begin{figure*}[t]
  \centering
   \includegraphics[width=1.0\linewidth]{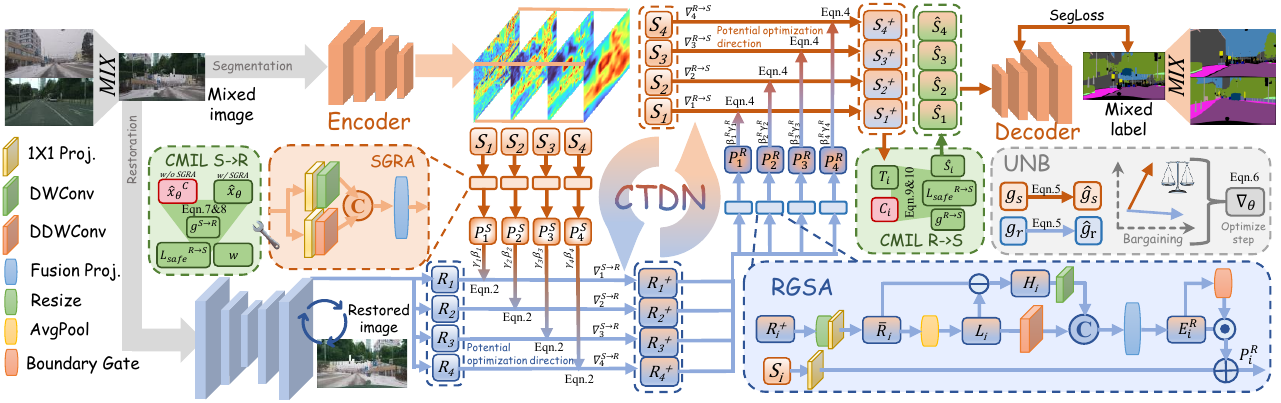}
   \caption{\textbf{Overview of the proposed Unsupervised Restoration-Segmentation Collaborative Learning Framework (Ultra)}. Given mixed images from cross-domain sampling, Ultra jointly optimizes image restoration and semantic segmentation through bidirectional cross-task guidance. Specifically, segmentation features $S_i$ are transformed into semantic priors $P_i^{S}$ by the SGRA, which performs conditional modulation on restoration features to provide potential restoration optimization directions. The modulated restoration features $R_i^{+}$ are further converted into restoration priors $P_i^{R}$ through the RGSA, providing potential segmentation optimization directions via conditional modulation. UNB selects the optimal cross-task optimization directions during training. To mitigate the Self-reinforcing Hallucination Loop in unsupervised cross-task collaboration, CMIL estimates the causal effects of semantic and restoration interventions and only propagates beneficial interventions.
   }
   \label{fig:framework}
\end{figure*}

In this paper, we propose an \textbf{Unsupervised Restoration-Segmentation Collaborative Learning Framework (Ultra)}, as shown in Figure~\ref{fig:framework}. During unsupervised domain adaptation, cross-domain mixed samples are jointly optimized for unsupervised image restoration and semantic segmentation. We propose CTDN, which enables mutual guidance between semantic segmentation and image restoration to generate candidate optimization directions and incorporates UNB to select the optimal cross-task direction. Finally, to address the Self-reinforcing Hallucination Loop, we design CMIL to estimate the causal effects of cross-task features, thereby preventing hallucination propagation.

\subsection{Cross-Task Direction Negotiation (CTDN)}
\label{sec:bima}
The segmentation branch provides category-aware spatial layouts and semantic structures that are complementary to the low-level appearance cues exploited by the diffusion restoration model. However, directly injecting segmentation features may introduce unreliable pseudo-semantic information under the unsupervised target domain. To this end, we propose a Segmentation-Guided Restoration Adapter (SGRA), which transforms segmentation representations into reliability-aware spatial-channel modulation signals for the diffusion denoising process. In detail, given the $i$-th stage segmentation feature $S_i$, we first obtain the semantic prior:
\begin{equation}
\resizebox{0.9\columnwidth}{!}{$
P_i^{S}
=
\psi_i
\left(
\left[
\mathrm{DWConv}_{3\times3}
(\phi_i(S_i))
\Vert
\mathrm{DDWConv}_{3\times3}
(\phi_i(S_i))
\right]
\right),
$}
\label{eq:semantic_prior}
\end{equation}
where $\phi_i(\cdot)$ denotes a $1\times1$ projection layer for channel
alignment, $\mathrm{DWConv}_{3\times3}(\cdot)$ and
$\mathrm{DDWConv}_{3\times3}(\cdot)$ denotes a $3\times3$ Depth-wise Convolution and a $3\times3$ Dilated Depth-wise Convolution, respectively, and $\psi_i(\cdot)$ denotes a
lightweight fusion projection layer. $\Vert$ means connection operation.

After semantic prior adaptation, a semantic prior modulation projection module is used to generate spatially varying scaling ($\gamma_i$) and shifting terms ($\beta_i$) from $P_i^{S}$. Then, conditional modulation is performed on the prior restoration feature as follows:
\begin{equation}
\resizebox{0.9\columnwidth}{!}{$
R_i^{+}
=
R_i+
\eta_i^{S\rightarrow R}
\left(
R_i\odot
(1+\lambda\tanh(\gamma_i))
+
\lambda\tanh(\beta_i)
-
R_i
\right),
$}
\label{eq:semantic_modulation}
\end{equation}
where $\lambda$ controls the modulation magnitude,
$\eta_i^{S\rightarrow R}$ is a zero-initialized residual coefficient. Then, the segmentation branch provides a potential optimization direction for the restoration branch at the $i$-th stage, denoted as $\Delta_i^{S\rightarrow R} = R_i^{+} - R_i$.

The restoration branch reconstructs fine-grained structural details and illumination-invariant contents that are complementary to the category-level semantics exploited by the segmentation model. However, directly injecting restoration features may transfer low-level appearance statistics, such as color and illumination, into the segmentation classifier. To address this issue, we propose a Restoration-Guided Segmentation Adapter (RGSA), which transforms restoration representations into boundary-aware spatial-channel modulation signals for the segmentation features.

Given the $i$-th stage restoration feature $R_i^{+}$ and segmentation feature $S_i$, we first obtain restoration prior ($P_i^{R}$): 
\begin{equation}
\resizebox{0.87\columnwidth}{!}{$
\left\{
\begin{array}{l}
\begin{aligned}
&
\bar{R}_i
=
\phi_i
\left(
\operatorname{Resize}(R_i^{+},S_i)
\right),
\\[3pt]
&
L_i
=
\operatorname{AvgPool}_{3\times3}(\bar{R}_i),
\qquad
H_i
=
\bar{R}_i-L_i,
\\[3pt]
&
E_i^{R}
=
\psi_i
\left(
\left[
\operatorname{DWConv}_{3\times3}(H_i)
\Vert
\operatorname{DDWConv}_{3\times3}(L_i)
\right]
\right),
\\[3pt]
&
P_i^{R}
=
\sigma
\left(
\kappa_i(E_i^{R})
\right)
\odot E_i^{R}
+
\phi_i(S_i).
\end{aligned}
\end{array}
\right.
$}
\label{eq:restoration_prior}
\end{equation}
where the symbols follow Eq.~\ref{eq:semantic_prior}. Then, $\operatorname{Resize}(R_i^{+},S_i)$ means bilinearly resizing $R_i^{+}$ to the spatial resolution of $S_i$. $L_i$ and $H_i$ are the low- and high-frequency components, respectively. 
$\sigma(\kappa_i(\cdot))$ is a depthwise boundary gate that suppresses
appearance-related responses.

After restoration prior adaptation, a structure-conditioned modulation projection module is used to generate structure-conditioned scaling ($\gamma_i^{R}$) and shifting terms ($\beta_i^{R}$) from $P_i^{R}$. Then, conditional modulation is performed on the prior segmentation feature: 
\begin{equation}
\resizebox{0.9\columnwidth}{!}{$
\begin{aligned}
S_i^{+}
&=
S_i
+
\eta_i^{R\rightarrow S}
\left[
S_i
\odot
\left(
1+\lambda\tanh(\gamma_i^{R})
\right)
+
\lambda\tanh(\beta_i^{R})
-
S_i
\right],
\end{aligned}
$}
\label{eq:restoration_modulation}
\end{equation}
where the coefficient $\eta_i^{R\rightarrow S}$ is a learnable residual
scale initialized to zero, ensuring that RGSA initially behaves as an identity mapping and gradually introduces restoration guidance during training.
Accordingly, the restoration branch provides a potential optimization
direction for the segmentation branch at stage $i$, defined as $\Delta_i^{R\rightarrow S}=S_i^{+}-S_i$.

\noindent\textbf{Unsupervised Nash Bargaining (UNB)}
To resolve this optimization conflict between unsupervised restoration and segmentation, we propose \textbf{UNB} to accumulate the detached segmentation gradients $\mathbf{g}_s$ and restoration gradients $\mathbf{g}_r$ at every backward pass, reweight them by the inverse of their EMA-smoothed task losses as a reliability estimate, and normalize them as follows:
\begin{equation}
\resizebox{0.8\columnwidth}{!}{$
\left\{
\begin{array}{l}
\begin{aligned}
&
\rho_k
=
\sqrt{
\operatorname{clamp}\!\left(
\frac{1/\bar{\mathcal{L}}_k}
{\frac{1}{2}\sum_{j\in\{s,r\}}1/\bar{\mathcal{L}}_j},
\;0.5,\,2
\right)
},
\\[5pt]
&
\hat{\mathbf{g}}_k
=
\frac{\rho_k\,\mathbf{g}_k}
{\|\rho_k\,\mathbf{g}_k\|_2}, \qquad
n_k
=
\|\rho_k\,\mathbf{g}_k\|_2,
\qquad
k\in\{s,r\}.
\end{aligned}
\end{array}
\right.
$}
\label{eq:unb_reliability}
\end{equation}
where $\bar{\mathcal{L}}_k$ denotes the EMA of the segmentation ($k{=}s$) or restoration ($k{=}r$) loss, $\rho_k$ is the detached reliability weight that up-weights the better-optimized task, and $n_k$ preserves the original gradient magnitude. The negotiated update is then the minimum-norm point on the convex hull of the two normalized task gradients, \textit{i.e.}, the Nash bargaining solution:
\begin{equation}
\resizebox{0.8\columnwidth}{!}{$
\left\{
\begin{array}{l}
\begin{aligned}
&
\alpha^{\ast}
=
\operatorname{clamp}\!\left(
\frac{G_{rr}-G_{sr}}
{G_{ss}+G_{rr}-2G_{sr}},
\;\omega,\,1-\omega
\right),
\\[5pt]
&
\nabla_{\theta_{\mathcal{A}}}
\leftarrow
\left(
\alpha^{\ast}\,\hat{\mathbf{g}}_s
+
(1-\alpha^{\ast})\,\hat{\mathbf{g}}_r
\right)
\cdot
\frac{n_s+n_r}{2}.
\end{aligned}
\end{array}
\right.
$}
\label{eq:unb_solution}
\end{equation} 
where $G_{ij}$ is the Gram inner product between the normalized task gradients, $\alpha^{\ast}$ is the bargaining coefficient clipped to $[\omega,1-\omega]$ to avoid degenerate single-task solutions, and the resulting direction overwrites the raw gradients on the shared parameters $\theta_{\mathcal{A}}$ before the optimizer step. Since $\alpha^{\ast}$ is derived from the gradient geometry rather than tuned by hand, the negotiated direction remains a valid descent direction for both objectives: when the two tasks cooperate ($G_{sr}>0$) the solution amplifies their consensus, and when they conflict ($G_{sr}<0$) it falls back to the direction that minimally violates either task.

\subsection{Causal Mutual Intervention Learning (CMIL)}
Considering that segmentation features from the unlabeled target domain inevitably contain pseudo-semantic noise under the unsupervised setting, blindly back-propagating restoration loss through such interventions may cause unreliable semantic gradients to dominate the restoration network. Therefore, we first estimate the causal effect of semantic interventions before gradient admission.

We use the restoration objective itself as the energy function and simulate one descent step for two parallel states. Let $\hat{x}_0$ and $\hat{x}_0^{c}$ denote the predictions generated with and without the SGRA intervention, respectively. Both states are detached from the training graph and optimized by one gradient step on the restoration energy $\mathcal{E}^{R}(\cdot)$. The causal effect of the semantic intervention $e^{S\rightarrow R}$ is then estimated by the counterfactual energy gap as follows: 
\begin{equation}
\resizebox{0.9\columnwidth}{!}{$
\begin{aligned}
e^{S\rightarrow R} = \mathrm{sg}\!\left[ \mathcal{E}^{R}(\hat{x}_0^{c} - \epsilon\,\nabla_{\hat{x}_0^{c}}\,\mathcal{E}^{R}(\hat{x}_0^{c})) - \mathcal{E}^{R}(\hat{x}_0 - \epsilon\,\nabla_{\hat{x}_0}\,\mathcal{E}^{R}(\hat{x}_0)) \right],
\end{aligned}
$}
\label{eq:s2r_effect}
\end{equation}
where $\epsilon$ is the simulation step size, $\nabla$ denotes the first-order gradient with respect to the detached state, $\mathrm{sg}[\cdot]$ is the stop-gradient operator. Intuitively, $e^{S\rightarrow R}>0$ indicates that the restoration energy descends faster after the semantic intervention, \textit{i.e.}, the direction $\Delta_i^{S\rightarrow R}$ provided by the segmentation branch is genuinely beneficial for restoration.

\begin{algorithm}[t] 
\caption{Algorithm of Ultra.} 
\label{alg:Framwork} 
\begin{algorithmic}[1] 
\STATE \textbf{Initialize:} Segmentation network $G_{\theta_S}$, restoration network $R_{\theta_R}$, UNB, CMIL module $\mathcal{C}$, shared parameters $\theta_{\mathcal{A}}$
\FOR{$t\gets$ 1 to $T$}
\STATE Generate $\Delta_i^{S\rightarrow R}=R_i^{+}-R_i$ \qquad 
$\triangleright$ Eq.~(\ref{eq:semantic_prior}), (\ref{eq:semantic_modulation})
\STATE Generate $\Delta_i^{R\rightarrow S}=S_i^{+}-S_i$ \qquad 
$\triangleright$ Eq.~(\ref{eq:restoration_prior}), (\ref{eq:restoration_modulation})
\STATE Select optimal optimization direction $\triangleright$ Eq.~(\ref{eq:unb_reliability}), (\ref{eq:unb_solution})
\STATE Causal intervention for semantic and restoration injection $e^{S\rightarrow R}$, $e^{R\rightarrow S}$ \qquad \qquad $\triangleright$ Eq.~(\ref{eq:s2r_effect}), (\ref{eq:r2s_effect})
\STATE Perform interventions \qquad \qquad $\triangleright$ Eq.~(\ref{eq:s2r_gate}), (\ref{eq:r2s_gate})  
\STATE Update $S_i$ with $\hat{S}_i= S_i + g^{R\rightarrow S}\,\Delta_i^{R\rightarrow S}$ 
\STATE Update $\theta_S,\theta_R,\theta_{\mathcal A}$ by minimizing $\mathcal{L}=
\mathcal{L}_{S}
+\lambda_R\mathcal{L}_{R}
+\lambda_C\mathcal{L}_{CMIL}$
\ENDFOR \\
\end{algorithmic}
\end{algorithm}

The estimated effect is converted into a bounded gate, which controls both a directional regularizer and the mixture between treated and control gradients as follows: 
\begin{equation}
\resizebox{0.9\columnwidth}{!}{$
\left\{
\begin{array}{l}
\begin{aligned}
&
g^{S\rightarrow R}
=
\mathrm{sg}\!\left[
\operatorname{ReLU}\!\left(
\tanh\!\left(
\frac{e^{S\rightarrow R}}
{\max(|e^{S\rightarrow R}|,\tau)}
\right)
\right)
\right],
\\[5pt]
&
\mathcal{L}_{safe}^{S\rightarrow R}
=
g^{S\rightarrow R}
\left(
1-
\cos\!\left(
\hat{x}_0-\hat{x}_0^{c},
-\nabla_{\hat{x}_0^{c}}\mathcal{E}^{R}(\hat{x}_0^{c})
\right)
\right)
+
\left(1-g^{S\rightarrow R}\right)
\left\|
\hat{x}_0-\hat{x}_0^{c}
\right\|_2^{2},
\\[5pt]
&
\mathcal{L}_{rest}
=
(\tfrac{1}{2}+\tfrac{1}{2}\,g^{S\rightarrow R}\,)
\mathcal{E}^{R}(\hat{x}_0)
+
(\tfrac{1}{2}-\tfrac{1}{2}\,g^{S\rightarrow R}\,)\,
\mathrm{sg}\!\left[
\mathcal{E}^{R}(\hat{x}_0^{c})
\right].
\end{aligned}
\end{array}
\right.
$}
\label{eq:s2r_gate}
\end{equation}
where $\tau$ is the effect temperature that normalizes the gate into $[0,1)$, $\cos(\cdot,\cdot)$ denotes cosine similarity between the image-space intervention $\hat{x}_0-\hat{x}_0^{c}$, \textit{i.e.}, the downstream manifestation of $\Delta_i^{S\rightarrow R}$, and the energy descent direction. Beneficial interventions ($g^{S\rightarrow R}\!\to\!1$) are encouraged to align with the descent direction and their gradients are fully admitted; harmful interventions ($g^{S\rightarrow R}\!\to\!0$) are pulled back in magnitude.

Symmetrically, the structural refinement $\Delta_i^{R\rightarrow S}$ produced by RGSA is examined before entering the segmentation loss, since restoration features may carry appearance statistics that are harmful to the classifier. The key difference from Eq.~\ref{eq:s2r_effect} is that no target-domain ground truth is available, so the review is conducted on an unsupervised segmentation energy $\mathcal{E}^{S}(\cdot)$, defined as the normalized prediction entropy plus the confidence-weighted cross-entropy against trusted pseudo-labels.

For each stage $i$, the factual feature $S_i$ and the refined candidate $S_i^{+}$ are detached into a control state $C_i=\mathrm{sg}[S_i]$ and a treated state $T_i=\mathrm{sg}[S_i+\Delta_i^{R\rightarrow S}]$. Each state is advanced by one simulated descent step on $\mathcal{E}^{S}(\cdot)$, and the causal effect of adopting the refinement is estimated over all $N$ stages as:
\begin{equation}
\resizebox{0.9\columnwidth}{!}{$
\begin{aligned}
e^{R\rightarrow S} = \mathrm{sg}\!\left[ \frac{1}{N} \sum_{i=1}^{N} \left( \mathcal{E}^{S}( C_i - \epsilon\,\nabla_{C_i}\,\mathcal{E}^{S}(\{C_i\})) - \mathcal{E}^{S}(T_i - \epsilon\,\nabla_{T_i}\,\mathcal{E}^{S}(\{T_i\})) \right) \right],
\end{aligned}
$}
\label{eq:r2s_effect}
\end{equation}
Intuitively, $e^{R\rightarrow S}>0$ answers the question ``what would happen at the next optimization step if the refinement were adopted'': the segmentation energy would descend faster, \textit{i.e.}, the restoration prior $P_i^{R}$ provides structural information that the segmentation branch itself cannot obtain from the unlabeled target domain.

The refinement is accepted in proportion to its estimated effect, while a safe loss shapes the live intervention produced by RGSA as follows:
\begin{equation}
\resizebox{0.86\columnwidth}{!}{$
\left\{
\begin{array}{l}
\begin{aligned}
&
g^{R\rightarrow S}
=
\mathrm{sg}\!\left[
\operatorname{ReLU}\!\left(
\tanh\!\left(
\frac{e^{R\rightarrow S}}
{\max(|e^{R\rightarrow S}|,\tau)}
\right)
\right)
\right],
\\[5pt]
&
\mathcal{L}_{safe}^{R\rightarrow S}
=
g^{R\rightarrow S}
\sum_{i=1}^{N}
\left(
1-
\cos\!\left(
\Delta_i^{R\rightarrow S},
-\nabla_{C_i}\mathcal{E}^{S}(\{C_i\})
\right)
\right)
+
\left(1-g^{R\rightarrow S}\right)
\sum_{i=1}^{N}
\left\|
\Delta_i^{R\rightarrow S}
\right\|_2^{2}.
\end{aligned}
\end{array}
\right.
$}
\label{eq:r2s_gate}
\end{equation}
where the symbols follow Eq.~\ref{eq:s2r_gate}.
Then, $\hat{S}_i= S_i + g^{R\rightarrow S}\,\Delta_i^{R\rightarrow S}$ is the accepted segmentation feature that replaces $S_i$ in the subsequent decoder. Since $g^{R\rightarrow S}$ is detached, the gate itself cannot become a learnable shortcut; instead, $\mathcal{L}_{safe}^{R\rightarrow S}$ pulls the live intervention $\Delta_i^{R\rightarrow S}$ toward the segmentation descent direction when the refinement is beneficial, and suppresses its magnitude back to zero otherwise. The final modulated features $\{\hat{S}_i\}$, together with the source-supervised loss, are then used to compute the segmentation objectives, and $\mathcal{L}_{safe}^{R\rightarrow S}$ is back-propagated with a small weight as an additional segmentation-side regularizer. The full training procedure is outlined in Algorithm 1.

\begin{table}
\renewcommand\arraystretch{1.2}
\resizebox{\linewidth}{!}{
\begin{tabular}{ll|cc}
\hline
Method & Backbone & ACDC test & ACDC val \\ \hline
DeepLabV2~\cite{chen2017deeplab}
& DeepLabV2 & 38.0 & 34.6 \\
Refign~\cite{bruggemann2022refign}
& DeepLabV2 & 48.0 & 55.8 \\
CMA~\cite{bruggemann2023contrastive}
& DeepLabV2 & 50.4 & 48.8 \\
VBLC~\cite{li2023vblc}
&DeepLabV2 & 47.8 & 46.0 \\
\rowcolor{Green} \textbf{Ultra (Ours)}
& DeepLabV2 & \textbf{58.6} & \textbf{56.7} \\ \hline

DAFormer~\cite{hoyer2022daformer}
& DAFormer & 55.3 & 55.3 \\
Refign~\cite{bruggemann2022refign}
& DAFormer & \textbf{65.5} & 65.0 \\
VBLC~\cite{li2023vblc}
& DAFormer & 64.2 & 63.7 \\
CoPT~\cite{mata2025copt}
& DAFormer & 63.7 & 64.8 \\
Instance-Warp~\cite{zheng2025instance}
& DAFormer & 61.7 & 61.8 \\
\rowcolor{Green} \textbf{Ultra (Ours)}
& DAFormer & \textbf{65.5} & \textbf{65.6} \\ \hline

HRDA~\cite{hoyer2022hrda}
& HRDA & 68.0 & 65.2 \\
Refign~\cite{bruggemann2022refign}
& HRDA & 72.1 & 71.1 \\
VBLC~\cite{li2023vblc}
& HRDA & 67.2 & 66.5 \\
CoDA~\cite{gong2024coda}
& HRDA & 72.6 & 72.4 \\
ACSegFormer~\cite{liu2024domain}
& HRDA & 72.7 & 72.6 \\
CISS~\cite{sakaridis2025condition}
& HRDA & 69.6 & 68.7 \\
\rowcolor{Green} \textbf{Ultra (Ours)}
& HRDA & \textbf{73.0} & \textbf{73.0} \\ \hline
\end{tabular}
}
\caption{Comparison of the state-of-the-art in Cityscapes$\rightarrow$ACDC domain adaptation on the ACDC test and ACDC val sets. The evaluation metric is mIoU, where a higher value indicates better performance. The best results are presented in \textbf{bold}. 
}
\label{t:acdctval}
\end{table}

\begin{figure*}[t]
  \centering
   \includegraphics[width=1.0\textwidth]{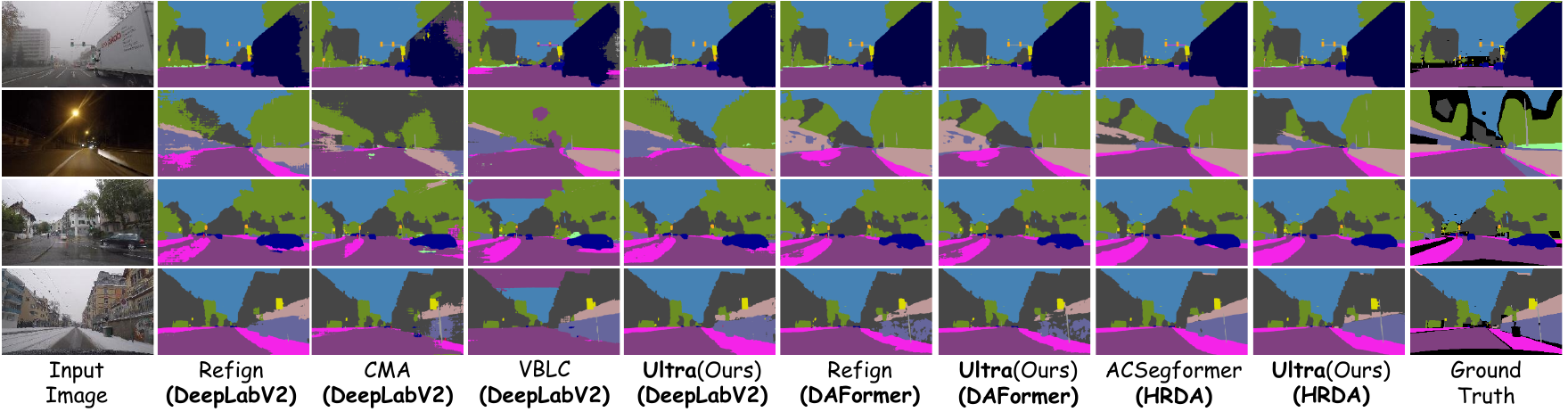}
   \caption{Qualitative comparison of our method and existing UDA-ASS approaches with DeepLabV2, DAFormer, and HRDA backbones on the ACDC val set. Under the same backbone configurations, our method generates predictions that better match the ground-truth annotations compared with previous approaches (\textit{e.g.}, Refign, CMA, VBLC, and ACSegFormer). The HRDA-based model achieves the best segmentation quality, further demonstrating the effectiveness of the proposed framework.}
   \label{fig:experiments}
\end{figure*}

\section{Experiments}
\label{sec:experiments}
We adopt mean Intersection-over-Union (mIoU) as the segmentation evaluation metric, where higher scores indicate superior segmentation performance. 
We assess our method under three settings: 1) UDA-ASS with Cityscapes$\rightarrow$ACDC adaptation; 2) UDA night-time semantic segmentation with Cityscapes$\rightarrow$Dark Zurich adaptation; and 3) image restoration performance in the UDA-ASS scenario. Furthermore, we investigate the generalization ability of our approach to UDA object detection, specifically on the BDD100K Clear$\rightarrow$ACDC Object Detection benchmark.

\subsection{Experimental settings}
Our framework is implemented in PyTorch and trained on a single NVIDIA 80\,GB A100 GPU. We evaluate our approach with several segmentation backbones, including DeepLabV2~\cite{chen2017deeplab}, DAFormer~\cite{hoyer2022daformer}, and HRDA~\cite{hoyer2022hrda}. The training process consists of 60k iterations, where 1024$\times$1024 image crops are randomly sampled from the Cityscapes and ACDC datasets. We optimize the model using AdamW with a weight decay of $1\times10^{-4}$. To alleviate the long-tail category imbalance in the source domain, we employ the rare class sampling strategy introduced in~\cite{hoyer2022daformer} with $\alpha=0.999$. 

\subsection{Comparison with State-of-the-art Methods}
\subsubsection{Comparison on ACDC}
We present comparisons to several kinds of semantic segmentation methods, including 1) \textit{backbones}: DeepLab-v2~\cite{chen2017deeplab}, DAFormer~\cite{hoyer2022daformer}, and HRDA~\cite{hoyer2022hrda}; 
2) \textit{UDA-SS methods under Adverse Weather}: Refign~\cite{bruggemann2022refign}, CMA~\cite{bruggemann2023contrastive}, VBLC~\cite{li2023vblc}, CoDA~\cite{gong2024coda}, and ACSegFormer~\cite{liu2024domain}; and 3) \textit{general UDA-SS methods}: CISS~\cite{sakaridis2025condition}, CoPT~\cite{mata2025copt} and Instance-Warp~\cite{zheng2025instance}; 
The quantitative results of mIoU performance on the ACDC test set and the ACDC val set are reported in Table~\ref{t:acdctval}. 
We observe that our method consistently outperforms existing approaches on the same backbone on the ACDC test and val sets. 
Notably, with HRDA as the backbone, our method achieves 73.0 [\%] mIoU on both the ACDC test and val sets, establishing state-of-the-art performance for Cityscapes $\rightarrow$ ACDC domain adaptation.

Figure~\ref{fig:experiments} provides qualitative comparisons between our method and existing UDA-ASS approaches on the ACDC val set. Under the same backbone setting, our method produces segmentation results that are more consistent with the ground-truth labels than competing methods. In particular, our method based on HRDA achieves the best performance, highlighting the effectiveness of the proposed framework.

\begin{figure}[t]
  \centering
   \includegraphics[width=\linewidth]{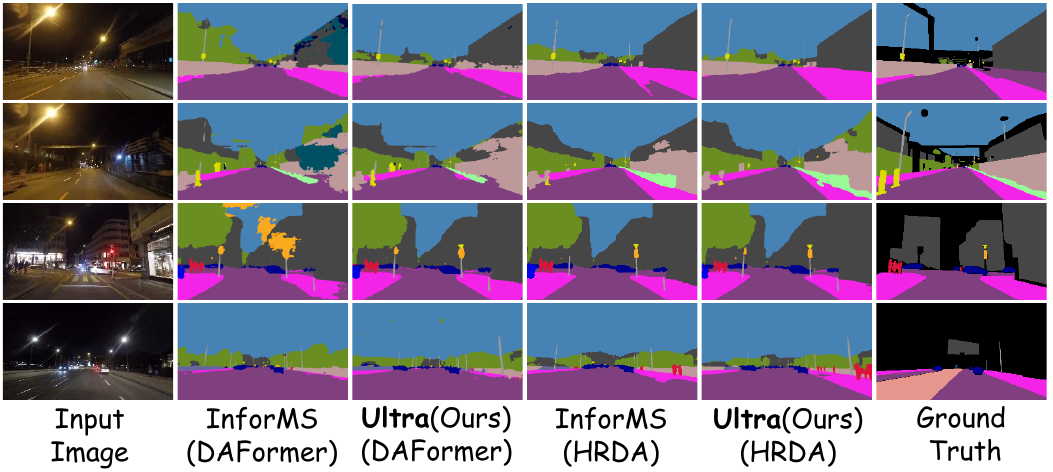}
   \caption{Qualitative comparison with existing UDA-ASS approaches using DAFormer and HRDA backbones on Dark Zurich val (top two rows) and Nighttime Driving test (bottom two rows).} 
   \label{fig:experiments_dark}
\end{figure}

\begin{table}
\renewcommand\arraystretch{1.2}
\resizebox{\linewidth}{!}{
\begin{tabular}{lccc}
\hline
\multirow{2}{*}{Method} & \multirow{2}{*}{Backbone}  & Dark Zurich-val & Nighttime Driving \\ \cline{3-4} 
                        &                            & mIoU            & mIoU          \\ \hline
DAFormer~\cite{hoyer2022daformer}            & DAFormer                    & 37.1               &  54.0          \\
InforMS~\cite{wang2023informative}      & DAFormer                   & 45.1          & 56.0    
\\ 
\rowcolor{Green} \textbf{Ultra (Ours)}         & DAFormer                  &   \textbf{45.8}        & \textbf{56.8}
\\ \hline
HRDA~\cite{hoyer2022hrda}            & HRDA                    &  42.1              & 54.1          \\
InforMS~\cite{wang2023informative}         & HRDA               & 52.5            & 58.5  \\
\rowcolor{Green} \textbf{Ultra (Ours)}         & HRDA                   &   \textbf{52.8}         & \textbf{59.3}   \\
\hline
\end{tabular}
}
\caption{Comparison with state-of-the-art methods on the Dark Zurich-val set and Nighttime Driving test set when performing Cityscapes$\rightarrow$Dark Zurich domain adaptation.}
\label{t:night}
\end{table}

\subsubsection{Comparison on Dark Zurich}
To evaluate the effectiveness of our framework under nighttime scenarios, we conduct experiments on the Cityscapes$\rightarrow$Dark Zurich domain adaptation setting and report the performance on the Dark Zurich val dataset in Table~\ref{t:night}.
Our HRDA-based model achieves 52.8 [\%] mIoU, surpassing previous approaches and establishing state-of-the-art performance. 
As shown in Figure~\ref{fig:experiments_dark}, our method produces more accurate and visually consistent predictions compared with existing methods, confirming its effectiveness in challenging nighttime environments.

\subsubsection{Comparison on Nighttime Driving}
To further examine the generalization ability of our framework under nighttime conditions, we evaluate the Cityscapes$\rightarrow$Dark Zurich adaptation model on the Nighttime Driving test set, with quantitative results reported in Table~\ref{t:night} and qualitative examples shown in Figure~\ref{fig:experiments_dark}. Using HRDA as the segmentation backbone, our method achieves the best performance on this benchmark, reaching 59.3 [\%] mIoU. These results demonstrate the robustness of our framework in handling challenging nighttime scenarios.

\subsubsection{Comparison with Image Restoration in UDA-ASS}
\begin{figure}[t]
  \centering
   \includegraphics[width=\linewidth]{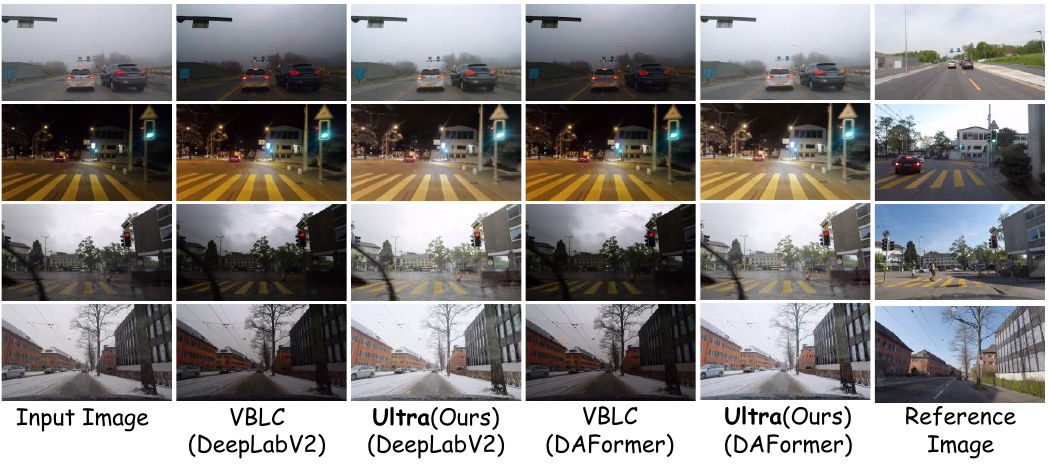}
   \caption{Qualitative comparison of Ultra (ours) and VBLC on ACDC val. Our method produces more natural restored images that are closer to the coarse-aligned target-domain reference images provided by ACDC~\cite{sakaridis2024acdcadverseconditionsdataset}.} 
   \label{fig:restoration}
\end{figure}

To validate the effectiveness of our unsupervised image restoration method, we compare it with the image restoration algorithms used in existing UDA-ASS methods, such as VBLC, which follows a restore-then-segment paradigm. The quantitative and qualitative results of unsupervised image restoration on the ACDC val set are reported in Table~\ref{t:restoration} and Figure~\ref{fig:restoration}, respectively. Different from conventional static restoration methods, our framework leverages cross-task optimization to achieve superior image restoration quality, further validating the effectiveness of the proposed approach.

\begin{table}
\renewcommand\arraystretch{1.2}
\centering
\resizebox{0.7\linewidth}{!}{
\begin{tabular}{lcc}
\hline
\multirow{2}{*}{Method}   & DeepLabV2 & DAFormer \\ \cline{2-3} 
                                           & LOE$\downarrow$            & LOE$\downarrow$         \\ \hline
VBLC~\cite{li2023vblc}                         &   37.90             &    37.90        \\
\rowcolor{Green} \textbf{Ultra (Ours)}                   &   \textbf{30.26}        &  \textbf{29.98} 
\\ \hline
\end{tabular}
}
\caption{Comparison of image restoration performance between our method and existing UDA-ASS restoration methods on the ACDC val set. The no-reference LOE metric is adopted, where lower values indicate better performance.}
\label{t:restoration}
\end{table}

\subsection{Ablation Study}
In this section, we validate the effectiveness of the three core components. For the Cityscapes $\rightarrow$ ACDC domain adaptation task, we train our model variants with different configurations and evaluate their performance on the ACDC val set, as shown in Table~\ref{t:ablation}. 
We first perform target-domain image restoration and semantic segmentation in parallel, and then gradually introduce the proposed components. 
The results show that adding CTDN without UNB improves mIoU from 65.0 to 65.2 on ACDC val, demonstrating the effectiveness of transforming visual structural priors and semantic discriminative information into mutually constrained optimization directions through bidirectional task adaptation. 
While enabling UNB further raises it to 65.4, indicating that the effectiveness of joint-benefit-based direction negotiation. 
Then, incorporating CMIL further improves the performance to 65.5. 
Finally, the full model achieves the best performance of 65.6 [\%] mIoU, verifying the complementary roles and necessity of each component in unsupervised restoration-segmentation collaborative adaptation. 

\begin{table}[]
\centering
\renewcommand\arraystretch{1.2}
\resizebox{1.0\linewidth}{!}{
\begin{tabular}{ccc|c}
\hline
\multicolumn{2}{c}{Cross-Task Direction Negotiation (CTDN)} & \multirow{2}{*}{Causal Mutual Intervention Learning (CMIL)}  & \multirow{2}{*}{ACDC val} \\ \cline{1-2}
w/o Unsupervised Nash Bargaining (UNB)        &  w/ UNB      &               &                             \\ \hline
            &            &                        &       65.0 (+0.0)               \\
\checkmark           &            &              &    65.2 (+0.2)                               \\
           & \checkmark           & \multicolumn{1}{l|}{} & \multicolumn{1}{c}{65.4 (+0.4)}      \\
\checkmark            &            & \multicolumn{1}{c|}{\checkmark} & \multicolumn{1}{c}{65.5 (+0.5)}      \\
            & \checkmark          & \checkmark                      & \multicolumn{1}{c}{\textbf{65.6 (+0.6)}}     \\ \hline
\end{tabular}}
\caption{Ablation Study on several model variants of our method on the ACDC val (Backbone: DAFormer).}
\label{t:ablation}
\end{table}

\subsection{Generalization Study}

\begin{table}[]
\centering
\renewcommand\arraystretch{1.2}
\resizebox{\linewidth}{!}{
\begin{tabular}{l|ccccc}
\hline
Method        & mAP$\uparrow$ & mAP50$\uparrow$ & mAP75$\uparrow$ & mAPs$\uparrow$ & mAPm$\uparrow$ \\ \hline
2PCNet~\cite{kennerley20232pcnet}        & 16.6     & 30.8      &  14.9       & 6.0     &  24.8    \\
\rowcolor{Green} 2PCNet + \textbf{Ultra (Ours)}                 & \textbf{18.9 (+2.3)}    & \textbf{34.2 (+3.4)}      & \textbf{19.3 (+4.4)}      & \textbf{7.7 (+1.7)}    & \textbf{27.5 (+2.7)}   \\ \hline
Instance-Warp~\cite{zheng2025instance} & 17.9    &  32.1       &  18.2       & 6.8     &  26.7      \\
\rowcolor{Green} Instance-Warp + \textbf{Ultra (Ours)}                 & \textbf{18.8 (+0.9)}    & \textbf{33.4 (+1.3)}      & \textbf{18.6 (+0.4)}      & \textbf{6.9 (+0.1)}    & \textbf{28.3 (+1.6)}   \\ \hline
\end{tabular}
}

\caption{Comparison of UDA object detection performance on BDD100K Clear $\rightarrow$ ACDC Object Detection on the ACDC object detection validation set. The better result within each baseline pair is shown in \textbf{bold}.}
\label{t:acdcobject}
\end{table}
To verify the generalization of our proposed method, we incorporate CTDN and CMIL into the original unsupervised domain adaptation object detection framework to achieve unsupervised restoration and object detection collaborative optimization.
Experimental results are summarized in Table~\ref{t:acdcobject}. Notably, our method consistently yields significant performance gains, confirming its superior generalization capability. 
\section{Conclusion}
\label{sec:conclusion}
In this work, we investigate two critical challenges in unsupervised restoration-segmentation mutual guidance: Ambiguous Cross-task Optimization Directions and Self-reinforcing Hallucination Loops.
To address these challenges, we propose Ultra, an unsupervised cross-task optimization framework that enables restoration and segmentation to collaboratively improve through reliable task interaction. Ultra introduces CTDN to discover cooperative optimization trajectories by exploiting complementary visual restoration cues and semantic information, and further incorporates CMIL to regulate cross-task knowledge exchange through task-level intervention effects. Extensive experiments on multiple UDA-ASS benchmarks demonstrate the effectiveness of Ultra across different architectures, while its extension to unsupervised restoration and object detection collaboration tasks verifies its generalization. Our work provides a new perspective for unsupervised complementary task learning, where heterogeneous visual tasks can collaboratively optimize their objectives while maintaining reliable information interaction.
{
    \small
    \bibliographystyle{ieeenat_fullname}
    \bibliography{main}
}


\end{document}